\documentclass[11pt]{article}

\usepackage[final]{acl}

\usepackage{times}
\usepackage{latexsym}
\usepackage{tabularx}
\usepackage{algorithm}
\usepackage{algorithmic}
\usepackage{amsmath}
\usepackage{hhline}
\usepackage{booktabs}
\usepackage{multirow}
\usepackage{float}
\usepackage{array}
\usepackage{xcolor}
\usepackage{xurl}

\usepackage[T1]{fontenc}

\usepackage[utf8]{inputenc}

\usepackage{microtype}

\usepackage{inconsolata}

\usepackage{graphicx}

\title{LexRel: Benchmarking Legal Relation Extraction for Chinese Civil Cases}
\author{
  \textbf{Yida Cai\textsuperscript{1,2}}, 
  \textbf{Ranjuexiao Hu\textsuperscript{3}}, 
  \textbf{Huiyuan Xie\textsuperscript{1}\thanks{Corresponding author.}}, 
  \textbf{Chenyang Li\textsuperscript{4,5}},
  \textbf{Yun Liu\textsuperscript{1}},\\
  \textbf{Yuxiao Ye\textsuperscript{1}},
  \textbf{Zhenghao Liu\textsuperscript{6}}, 
  \textbf{Weixing Shen\textsuperscript{1}},
  \textbf{Zhiyuan Liu\textsuperscript{1}\footnotemark[1]}\\
  \textsuperscript{1}Tsinghua University
  \textsuperscript{2}Peking University
  \textsuperscript{3}University of Glasgow \\
  \textsuperscript{4}Beijing University of Posts and Telecommunications \\
  \textsuperscript{5}Queen Mary University of London
  \textsuperscript{6}Northeastern University \\
  \texttt{yidacai1130@gmail.com, \{xieh,liuzy\}@tsinghua.edu.cn}
}

\begin{document}
    
\maketitle
\begin{abstract}
Legal relations serve as an important analytical framework for dispute resolution in civil cases. However, legal relations in Chinese civil cases remain underexplored in the field of legal AI, largely due to the absence of comprehensive schemas. In this work, we first introduce a comprehensive schema for legal relations in civil cases, which contains a hierarchical taxonomy and definitions of arguments. Based on this schema, we formulate a legal relation extraction task and present \textbf{LexRel}, an expert-annotated benchmark for legal relation extraction in the Chinese civil law domain. We use \textbf{LexRel} to evaluate state-of-the-art large language models (LLMs) on legal relation extraction, showing that current LLMs exhibit significant limitations in accurately identifying civil legal relations. Furthermore, we demonstrate that explicitly incorporating information about legal relations leads to promising performance gains on other downstream legal AI tasks.\footnote{Data and code are available at \url{https://github.com/thunlp/LexRel}.}
\end{abstract}


\begin{figure*}[h!]
\centering
\includegraphics[width=\textwidth]{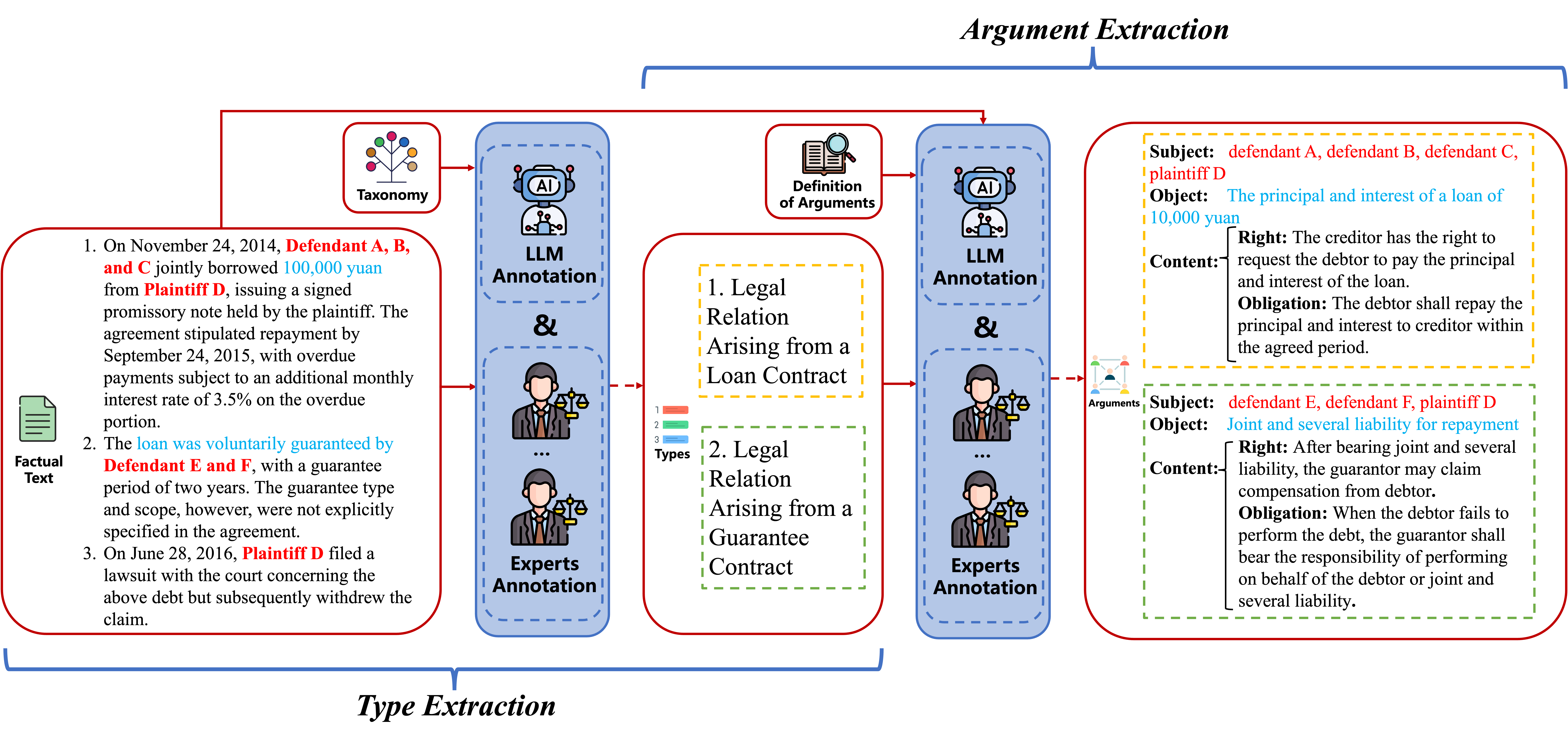} 
\caption{The construction workflow of \textbf{LexRel} is shown (English translation of the original Chinese text). The \textit{type extraction} task involves extracting types from factual text by referencing taxonomy, while the \textit{argument extraction} task involves extracting arguments from factual text and types by referencing definition of arguments. The \textcolor[rgb]{1, 0, 0}{red} and \textcolor[rgb]{0, 0.7, 0.94}{blue} annotations denote subject and object, as well as information that may represent them within the factual text.}
\label{fig1}
\end{figure*}

\section{Introduction}
Legal relations constitute a foundational analytical unit in Chinese civil cases. Defined as the relationships between individuals as regulated by legal norms, legal relations serve as the crucial interface between the normative world of law and the empirical world of social facts \cite{yuan2024reflections}. 
In judicial practice, legal professionals frequently rely on legal relations as a key reference for several tasks, such as legal information retrieval, statute prediction and case outcome analysis \cite{shen2004reconsideration}. However, while legal relations are central to both legal theory and practice, they remain underexplored in the field of legal AI, particularly in the context of Chinese civil law.
Legal information extraction in legal AI primarily targets factual entities (e.g., persons, goods, contracts) or general social relations (e.g., employment, ownership)~\cite{cui2024chatlawmultiagentcollaborativelegal,huang2021case,info15110666,li2021textguidedlegalknowledgegraph,xu2021mining}. Such approaches overlook that legal relations\textemdash a concept grounded in statutory rules and judicial practice\textemdash are qualitatively distinct from ordinary semantic associations in natural language. 
Moreover, in real-world practice, legal relations are seldom explicitly stated in judgments or other legal documents. As a result, models trained solely on large-scale legal texts often struggle to identify and apply legal relations accurately. Furthermore, existing legal relation schemas are typically coarse-grained, classifying relations at the broad level of civil rights and obligations~\cite{wan2022civil}. To enable AI systems to better distinguish and model legal relations, developing a structured, fine-grained schema of legal relations is therefore essential.

In this paper, we present the first comprehensive schema to model legal relations in Chinese civil law. Building upon established legal theory, real-world judicial practice, and expert legal guidance, our schema consists of a hierarchical taxonomy spanning 9 major domains of civil law and 265 relation types, together with precise definitions of their core arguments (subject, object and content). 

Based on this schema, we define the task of \textit{legal relation extraction}, which involves extracting legal relations from factual text. Building on this task, we develop \textbf{LexRel}, a benchmark consisting of 1,140 expert-annotated legal relations extracted from real-world cases. 

We use \textbf{LexRel} to evaluate state-of-the-art large language models (LLMs) on the task of \textit{legal relation extraction}. 
Our evaluation consists of two settings: a zero-shot baseline and a relation-enhanced baseline.
In addition, we apply the relation-enhanced approach to other legal AI tasks, demonstrating that integrating structured legal relation knowledge enhances downstream task performance.

Our contributions are summarized as follows:

\begin{itemize}
    \item We propose the first structured and comprehensive schema of legal relations in Chinese civil law, with a taxonomy covering 265 relation types across 9 major domains, and clear definitions of subjects, objects, and content.

    \item We propose the task of \textit{legal relation extraction}, and construct \textbf{LexRel}, an expert-annotated benchmark for \textit{legal relation extraction} in Chinese civil cases.

    \item We evaluate existing LLMs on \textbf{LexRel}, revealing significant model limitations in extracting legal relations from case facts.

    \item We demonstrate that integrating legal relations improves performance on downstream legal AI tasks.
\end{itemize}

\section{Related Work}
\subsection{Legal Relation}
Legal relations refer to social relations recognized and regulated by law, defined by the rights and obligations arising between parties by operation of law \cite{liang2021general}. As used in this work, the term primarily denotes civil legal relations under the Civil Code of the People's Republic of China~\cite{wan2022civil}. Within the new Civil Code, legal relations structure the organization of all chapters and form the basis for the logical framework of civil law~\cite{wang2018civil}. In judicial practice, legal relations also serve as a highly consequential methodological approach: judges situate the parties within a specific relation, analyze the subjects, objects, and their respective rights and obligations, and determine the creation, alteration, or extinction of rights~\cite{wang2017general}.

Beyond their doctrinal and practical roles, structured representations of legal relations can also benefit legal AI tasks. Prior work has shown that event- or relation-based representations can improve tasks such as case retrieval and legal summarization by enabling more precise fact matching and more coherent content planning~\cite{joshi2023ucreatunsupervisedcaseretrieval,santosh2025coperlexcontentplanningeventbased}. However, a structured and comprehensive schema of civil legal relations that can be directly operationalized for computational modeling has not yet been established.

\subsection{Legal Knowledge Graph} 
Legal knowledge graphs (KGs) are widely used in legal AI, powering applications such as legal reasoning~\cite{luo2024reasoninggraphsfaithfulinterpretable,li2021textguidedlegalknowledgegraph}, question answering~\cite{liang2024kagboostingllmsprofessional}, and decision support~\cite{cui2024chatlawmultiagentcollaborativelegal}. These graphs typically organize entities (e.g., persons, organizations, statutes) and their relationships into structured formats to enable downstream machine reasoning. 
However, most existing legal KGs in Chinese civil law focus on general social relations (e.g., employment, kinship, transaction) \cite{cui2024chatlawmultiagentcollaborativelegal,li2021textguidedlegalknowledgegraph} or entity co-occurrence patterns \cite{info15110666}.
That is, while legal KGs often capture who interacts with whom, they rarely characterize how they are related under law, such as the presence of a creditor-debtor relationship or a contractual obligation. 
Some works \cite{xu2021mining} adopt stricter definitions of civil legal relations to construct relation types, but their coverage remains limited and lacks a comprehensive taxonomy.

\subsection{Legal Large Language Models} 
Large language models (LLMs) show remarkable promise in legal applications, including legal consultation, trial assistance, and document analysis \cite{lai2023largelanguagemodelslaw}. Several works explore adapting general LLMs to legal domain through continued pretraining and fine-tuning \cite{huang2023lawyerllamatechnicalreport,xiao2021lawformer,yue2023disc}. In addition, some works focus on evaluating capabilities of LLMs in the legal field \cite{dai2023laiw,li2024lexeval}. These advancements demonstrate the growing capability of legal LLMs and highlight their strong potential to support a wide range of legal applications. 

\section{Schema for Civil Legal Relations}
\label{section3}
We present the first comprehensive schema for legal relations in Chinese Civil Law, which comprises a hierarchical taxonomy of relation types and precise definitions of their arguments.

\subsection{Taxonomy for Civil Legal Relations}
We construct the taxonomy of civil legal relations through a combination of initial taxonomy drafting and expert-guided refinement:

\begin{itemize}
    \item \textbf{Draft taxonomy construction.} We begin by applying a keyword-matching approach to extract relational expressions from Chinese civil court judgments~\cite{cjo2013}, producing an initial list of candidate legal relation terms. These candidates are then refined by referencing specialized legal terminology references~\cite{zhonghua1996}, retaining only those whose prefixes correspond to legally recognized concepts. 
    This process yields a draft taxonomy comprising 123 candidate relation types, categorized into six major domains based on the rights–obligations framework of Chinese civil law~\cite{wan2022civil}: \textit{Legal Relations of Personality Rights}, \textit{Legal Relations of Identity Rights}, \textit{Legal Relations of Intellectual Property Rights}, \textit{Legal Relations of Succession Rights}, \textit{Legal Relations of Property Rights}, and \textit{Legal Relations of Obligations}.

    \item \textbf{Expert-guided refinement.} Two senior legal scholars (law professors or lecturers) review and refine the draft taxonomy. Drawing on legal textbooks, academic literature, and their professional expertise, they introduce three new domains\textemdash \textit{Bill Relations}, \textit{Letter of Credit Relations}, and \textit{Independent Guarantee Relations}\textemdash expand the taxonomy to 265 legal relation types. 
\end{itemize}

These two stages, combining empirical induction from litigated cases with normative grounding in established legal doctrine, ensure that the resulting taxonomy is both data-informed and conceptually rigorous. The finalized taxonomy covers nine broad domains in Chinese civil law, comprising 265 fine-grained legal relation types. 

However, the distribution of relation types across these domains is not uniform.
Among them, \textit{Legal Relations of Obligations} encompasses the largest number of relation types, with 200 in total. 
This predominance reflects the inclusion of two categories corresponding to the most common types of civil disputes in practice\textemdash \textit{Contractual Legal Relations} and \textit{Legal Relations of Tort Liability}\textemdash which account for the majority of cases in judicial proceedings~\cite{supremecourt_china_2024}.

\subsection{Arguments of Civil Legal Relations}

\begin{table*}[t]
\centering
\fontsize{9}{11}\selectfont
\begin{tabularx}{\textwidth}{>{\hsize=0.15\hsize}X|>{\hsize=0.85\hsize}X}
    \toprule
    \multicolumn{2}{c}{\textbf{Definitions of Arguments in Legal Relations of Property Rights}} \\
    \toprule
    \textbf{Subject} & The right holder, together with any other subjects involved in the rights-and-obligations relationship, are all considered obligors. In practice, this includes the right holder himself/herself and any other parties who directly bear legal obligations under the relevant facts. \\
    \midrule
    \textbf{Object} & Corporeal things, including real property, movable property and special items. \\
    \midrule
    \textbf{Content} & In terms of rights, the right-holder legally enjoys ownership, usufructuary right, security interest, etc. The right-holder exercises direct control over the object through powers such as possess, use, seek profits from and dispose of the real property or movable property according to law. Real property rights become effective upon registration, whereas rights in movable property become effective upon delivery. In terms of obligations, it includes completing the registration or delivery procedures in accordance with the law; not infringing upon the legitimate property rights of others; and assuming the protection obligations such as removing obstacles, restoring the original state, and returning the original object when the property rights are unlawfully interfered with. \\
    \bottomrule
\end{tabularx}
\caption{An example of the definitions of the subject, object, and content for \textit{Legal Relations of Property Rights} (English translation for the original Chinese text). 
}
\label{table1}
\end{table*}

Having established the taxonomy of relation types, we next define their core arguments: \textit{subject}, \textit{object}, and \textit{content}~\cite{wang2018civil}.
In Chinese civil law, there are no precise definitions for different arguments in different types of legal relation.
With the help of the specific provisions of the Civil Code of the People's Republic of China regarding different categories of rights~\cite{wan2022civil} and senior legal scholars, we design precise definitions of the legal relations contained in main domains of our taxonomy.

Specifically, we select 8 of the 9 domains in the taxonomy, as well as 4 major subdomains under the domain \textit{Legal Relations of Obligations} to standardize 12 distinct definitions. The 4 subdomains are: \textit{Contractual Legal Relations}, \textit{Legal Relations of Tort Liability}, \textit{Legal Relations of Negotiorum Gestio} and \textit{Legal Relations of Unjust Enrichment}. All definitions follow the following coarse-grained definitions ~\cite{wang2018civil}:

\begin{itemize}
    \item \textit{Subject} refers to the party initiating or holding the right or obligation, typically a natural person, legal entity, or organization.
    \item \textit{Object} denotes the counterparty or legal target of the relation, such as another individual, a piece of property, or an intellectual asset.
    \item \textit{Content} denotes a paragraph that captures substantive nature of the legal relation, such as ownership, liability, or contractual duty.
\end{itemize}

Table \ref{table1} shows an example of definitions of arguments in a legal relation (for the complete set of definitions, refer to Appendix \ref{appendix:c}). 
As a result, each relation type in our taxonomy is associated with a specific configuration of these triplet elements, forming a unified schema that serves as the foundation for dataset construction and model evaluation in this work. 

\section{LexRel}

As shown in Fig.~\ref{fig1}, with the structured legal relation schema in place, we first define the task of extracting explicit legal relations from legal text that may contain implicit legal relation information as \textit{legal relation extraction}.
Based on this task definition, we construct \textbf{LexRel} benchmark. Each entry in \textbf{LexRel} consists of three components: factual text containing implicit legal relation information, a list of legal relation types, and the corresponding arguments (subject, object, and content) for each legal relation. 
\textbf{LexRel} is carefully curated from real-world Chinese civil court decisions through a two-step process: first, use LLMs to annotate candidate relations and their corresponding arguments, and second, legal experts review and refine the annotations to ensure both coverage and correctness.

\subsection{Task Definition}

\begin{table*}[h]
\centering
\fontsize{9}{10}\selectfont
\begin{tabularx}{\textwidth}{>{\hsize=0.5\hsize}X|>{\hsize=0.5\hsize}X}
    \toprule
    \textbf{Type Extraction} & \textbf{Argument Extraction}\\
    \toprule
    Please extract the existing legal relations from the following judgment text. Only output the names, no need for explanation or description. 
    
    \textbf{Factual text: \{\}}
    
    \textbf{Legal relation candidate set: \{\}}
    
    Output requirements:  1. Based on the content of the judgment, find the names of the legal relationships existing in the judgment text.  2. Only output the name of the most important legal relationship existing in the judgment text, and the output relationship name should be within 10 characters.  3. Each legal relationship name output must be in the legal relation candidate set.  4. Output strictly in accordance with the output format, and the output format is: [Legal relation 1, Legal relation 2......] 
    &  Please extract the subject, object, and content of \textbf{\{relation type\}}  from the given judgment documents without explanation or description.

    \textbf{Factual text: \{\}}.
    
    \textbf{Subject definition: \{\}}.

    \textbf{Object definition: \{\}}.
    
    \textbf{Content definition: \{\}}.
    
    Output requirements:
    1. Each set of results is presented as a standard JSON object.
    2. Required fields: \{`subject': `', `object': `', `content': `'\}.
    3. Each set is in a separate paragraph without numbering or sorting.
    4. Only one set needs to be output for the same subject and object. \\
    \bottomrule
\end{tabularx}
\caption{Prompt templates used in \textit{type extraction} and \textit{argument extraction} are shown (English translation for the original Chinese text). In \textit{type extraction}, the additional inputs required are factual text and the collection of all relation types from the taxonomy. In \textit{argument extraction}, supplementary inputs include factual text, one relation type from the set of previously extracted relation types, together with the definitions of three arguments.}
\label{table2}
\end{table*}
Specifically, \textit{legal relation extraction} comprises two subtasks: \textit{type extraction}, which aims to identify the relevant legal relation types, and \textit{argument extraction}, which seeks to determine the specific knowledge in each relation.

\subsubsection{Step 1: Type Extraction}

Given input case facts \( x \), which consists of the fact statements in a court judgment, the type extraction stage aims to extract the legal relation type \( \hat{r} \):
\begin{equation}
\hat{r} = f_{\text{type}}(x), \quad \hat{r} \in \mathcal{R}
\end{equation}

Here, \( f_{\text{type}} \) denotes relation type classifier, and \( \mathcal{R} \) is the legal relation set predefined in our taxonomy.

\subsubsection{Step 2: Argument Extraction}

Conditioned on the predicted relation type \( \hat{r} \), the model then extracts a set of associated subjects $\hat{S} = \{ \hat{s}_1, \hat{s}_2, \ldots \}$, a set of associated objects $\hat{O} = \{ \hat{o}_1, \hat{o}_2, \ldots \}$, and a detailed content (i.e., rights and obligations of the parties) $\hat{c}$. The extractor \( f_{\text{arg}} \) produces argument sets based on both input facts \( x \) and relation type \( \hat{r} \):

\begin{equation}
(\hat{S}, \hat{O}, \hat{c}) = f_{\text{arg}}(x, \hat{r})
\end{equation}

\subsection{Benchmark Construction}
To ensure the quality of extraction and cost-effectiveness, we adopt a pipeline in which extraction is first performed by an LLM and then corrected by experts.

\paragraph{Data preprocessing.}

We draw on court judgments for real-world adjudicated civil trial cases to construct the \textbf{LexRel}. First, we utilize DeepSeek-V3 \cite{deepseekai2025deepseekv3technicalreport} to extract legal relation types and their arguments from full-text judgment producing draft annotations to guide human experts during annotation. The extraction follows the two-step task structure defined above, with prompting instructions provided in Table \ref{table2}.

We then use DeepSeek-V3 to extract factual text recognized by the court from each judgment as the input to \textbf{LexRel}. The LLM is prompted to identify factual paragraphs that reflect adjudicated evidence, excluding legal analysis and procedural history. 

\paragraph{Expert annotation.} We engage six legal experts for annotation, all of whom hold law degrees and have prior experience in data annotation. They are provided with detailed guidelines describing  our schema and are briefed to ensure consistent understanding. Each annotator is assigned 200 items, each containing the facts and court reasoning extracted from judgments, together with draft relation types and arguments. They are tasked with verifying and refining input facts, types and arguments. 
The annotation process is overseen by a senior expert in legal AI research with extensive knowledge of legal relations. An analysis of annotators' agreement is provided in Appendix \ref{appendix:d}. 
After annotation, 60 items without identifiable legal relations are removed. The final \textbf{LexRel} benchmark comprises 1,140 annotated samples, each consisting of factual text paired with legal relations and their corresponding arguments, forming the \textbf{LexRel} evaluation dataset for assessing models' ability to extract legal relations from case facts.

\section{Experiment}
\subsection{Evaluation on LLMs}
We begin by using \textbf{LexRel} to evaluate state-of-the-art large language models (LLMs) on the task of \textit{legal relation extraction} under two settings: a zero-shot baseline and a relation-enhanced baseline. The zero-shot baseline assesses the models' ability to extract legal relations without any task-specific fine-tuning. In contrast, the relation-enhanced baseline involves providing LLMs with the full judgment text rather than only the factual text. This enables LLMs to generate relation types and arguments, which are subsequently used as supervision to train smaller models via supervised fine-tuning (SFT).

\begin{table*}[t]
  \centering
  \fontsize{7.5}{12.5}\selectfont
  \begin{tabular}{l|l|cccc|cccc}
    \toprule
    \multirow{2}{*}{\textbf{Models}} & \multirow{2}{*}{\textbf{Methods}} 
    & \multicolumn{4}{c|}{\textbf{Type Extraction}} 
    & \multicolumn{4}{c}{\textbf{Argument Extraction}} \\
    \cmidrule(lr){3-6} \cmidrule(lr){7-10}
    & & \textbf{P} & \textbf{R} & \textbf{micro-F1} & \textbf{macro-F1} 
      & \textbf{P} & \textbf{R} & \textbf{micro-F1} & \textbf{macro-F1} \\
    \midrule
    \textbf{GPT-4o}         & zero-shot & 0.582 & \textbf{0.789} & 0.670 & 0.314 & 0.188 & 0.278 & 0.224 & 0.068 \\
    \textbf{o3-mini}    & zero-shot & \textbf{0.768} & 0.756 & \textbf{0.762} & \textbf{0.441} & \textbf{0.413} & 0.355 & \textbf{0.382} & \textbf{0.129}\\
    \textbf{Claude-Sonnet-4} & zero-shot & 0.619 & 0.564 & 0.590 & 0.330 & 0.206 & 0.345 & 0.258 & 0.088\\ 
    \textbf{DeepSeek-R1}    & zero-shot & 0.647 & 0.746 & 0.693 & 0.376 & 0.299 & 0.242 & 0.268 & 0.065\\
    \textbf{DeepSeek-V3}    & zero-shot & 0.440 & 0.781 & 0.563 & 0.308 & 0.164 & 0.430 & 0.237 & 0.083\\
    \textbf{Llama3.1-70B}   & zero-shot & 0.192 & 0.384 & 0.256 & 0.074 & 0.022 & 0.203 & 0.039 & 0.008\\
    \textbf{Qwen3-32B}     & zero-shot & 0.500 & 0.700 & 0.583 &  0.158 & 0.030 & \textbf{0.551} & 0.057 & 0.015\\ 
    \midrule
    \midrule
    
    \multirow{3}{*}{\textbf{Llama3.1-8B-Instruct}}    & zero-shot & 0.170 & 0.472 & 0.250 & 0.052 & 0.014 & 0.337 & 0.027 & 0.006\\
    & RE \textit{w/ GPT-4o} & 0.488 &  0.741 & 0.588 & 0.270 & 0.131 & 0.287 & 0.180 & 0.054\\
    & RE \textit{w/ DeepSeek-R1} & 0.586 & 0.771 & 0.666 & 0.338 & 0.236 & 0.339 & 0.279 & 0.093\\
    \midrule
    
    \multirow{3}{*}{\textbf{InternLM3-8B-Instruct}}   & zero-shot & 0.310 & 0.644 & 0.418 & 0.139 & 0.055 & 0.042 & 0.048 & 0.002\\
    & RE \textit{w/ GPT-4o} & 0.479 &  0.773 & 0.592 & 0.288 & 0.150 & 0.349 & 0.209 & 0.061\\
    & RE \textit{w/ DeepSeek-R1} & 0.599 & \underline{0.817} & 0.691 & 0.324 & 0.277 & 0.387 & 0.323 & 0.100 \\
    \midrule
    
    \multirow{3}{*}{\textbf{MiniCPM4-8B}}    & zero-shot & 0.353 & 0.417 & 0.382 & 0.218 & 0.095 & 0.128 & 0.109 & 0.053 \\
    & RE \textit{w/ GPT-4o} & 0.519 &  0.760 & 0.617 & 0.314 & 0.162 & 0.314 & 0.213 & 0.054\\
    & RE \textit{w/ DeepSeek-R1} & 0.599 & 0.769 & 0.674 & 0.343 & 0.243 & 0.310 & 0.272 & 0.086\\
    \midrule
    
    \multirow{3}{*}{\textbf{Qwen3-8B}}       & zero-shot & 0.338 & 0.745 & 0.464 & 0.149 & 0.061 & 0.275 & 0.099 & 0.014\\
    & RE \textit{w/ GPT-4o} & 0.536 & 0.769  & 0.632 & 0.311 & 0.175 & 0.346 & 0.233 & 0.078\\ 
    & RE \textit{w/ DeepSeek-R1} & 0.576 & 0.816 & 0.675 & 0.337 & 0.255 & 0.375 & 0.304  & 0.098\\
    \midrule
    
    \multirow{3}{*}{\textbf{Qwen3-14B}}      & zero-shot & 0.564 & 0.603 & 0.583 & 0.313 & 0.072 & 0.173 & 0.102 & 0.035 \\
    & RE \textit{w/ GPT-4o} & 0.614 & 0.749 & 0.675 & 0.357 & 0.212 & 0.351 & 0.264 & 0.096\\
    & RE \textit{w/ DeepSeek-R1} & \underline{0.719} & 0.748 & \underline{0.733} & \underline{0.430} & \underline{0.373} & \underline{0.389} & \underline{0.381} & \underline{0.146}\\
    \bottomrule
  \end{tabular}
  \caption{Evaluation results of state-of-art LLMs on \textbf{LexRel} under precision (P), recall (R), micro-F1 and macro-F1 to \textit{type extraction} and \textit{argument extraction} are shown. The first 7 models are evaluated only on the zero-shot baseline due to the consideration of training costs, while the remaining models with 8B and 14B parameters are evaluated under both zero-shot and relation-enhanced (RE) baselines. The notation “\textit{w/}” specifies which model is used to generate the training data. Across all results, the best zero-shot performance for each metric is shown in \textbf{bold}, and the best performance under relation-enhanced baseline is \underline{underlined}.}
  \label{table3}
\end{table*}

\begin{figure*}[h]
\centering
\begin{minipage}{0.48\textwidth}
    \centering
    \includegraphics[width=\linewidth]{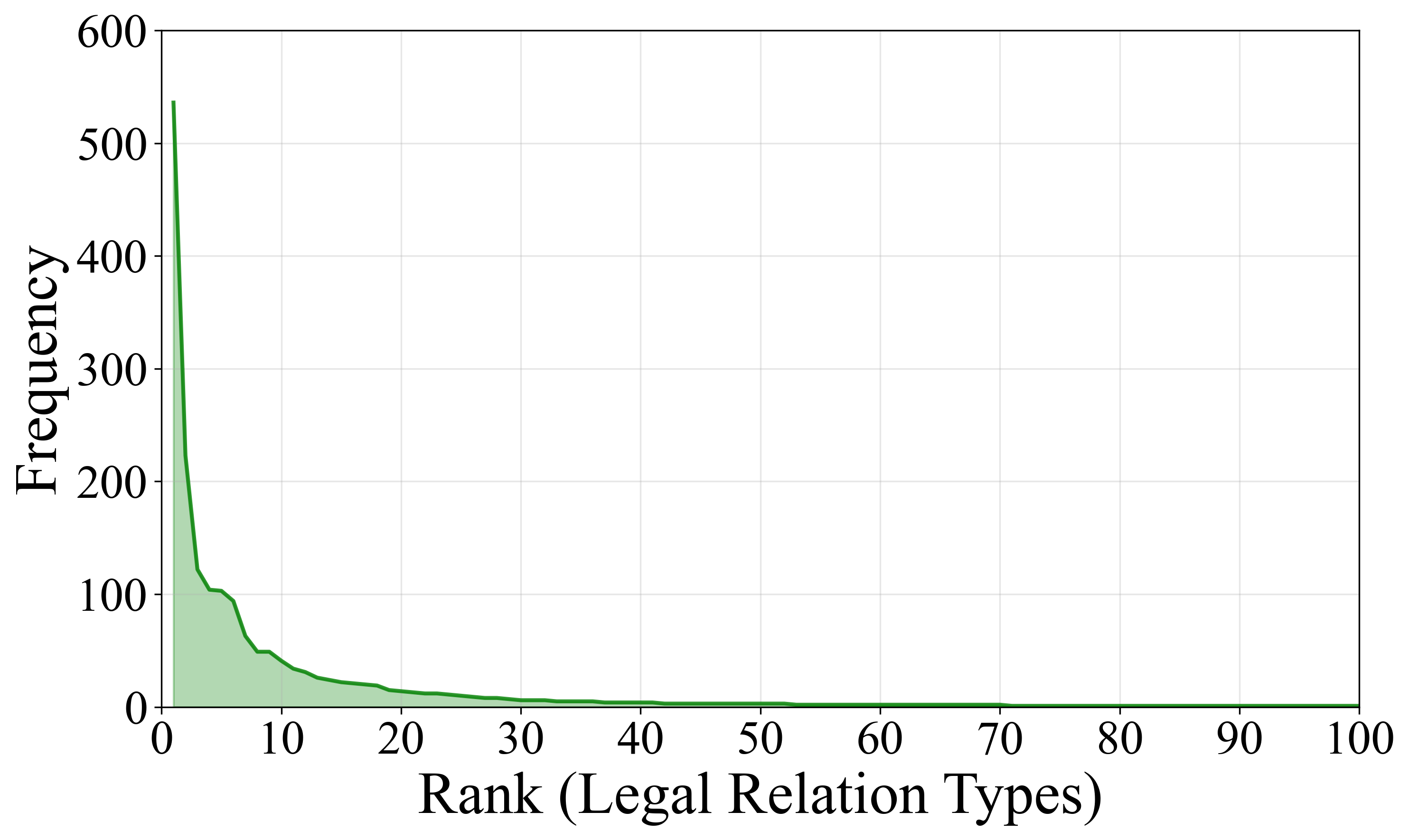}
    \caption*{(A) Distribution of top 100 most frequent legal relation types in \textbf{LexRel}.}
\end{minipage}
\hfill
\begin{minipage}{0.48\textwidth}
    \centering
    \includegraphics[width=\linewidth]{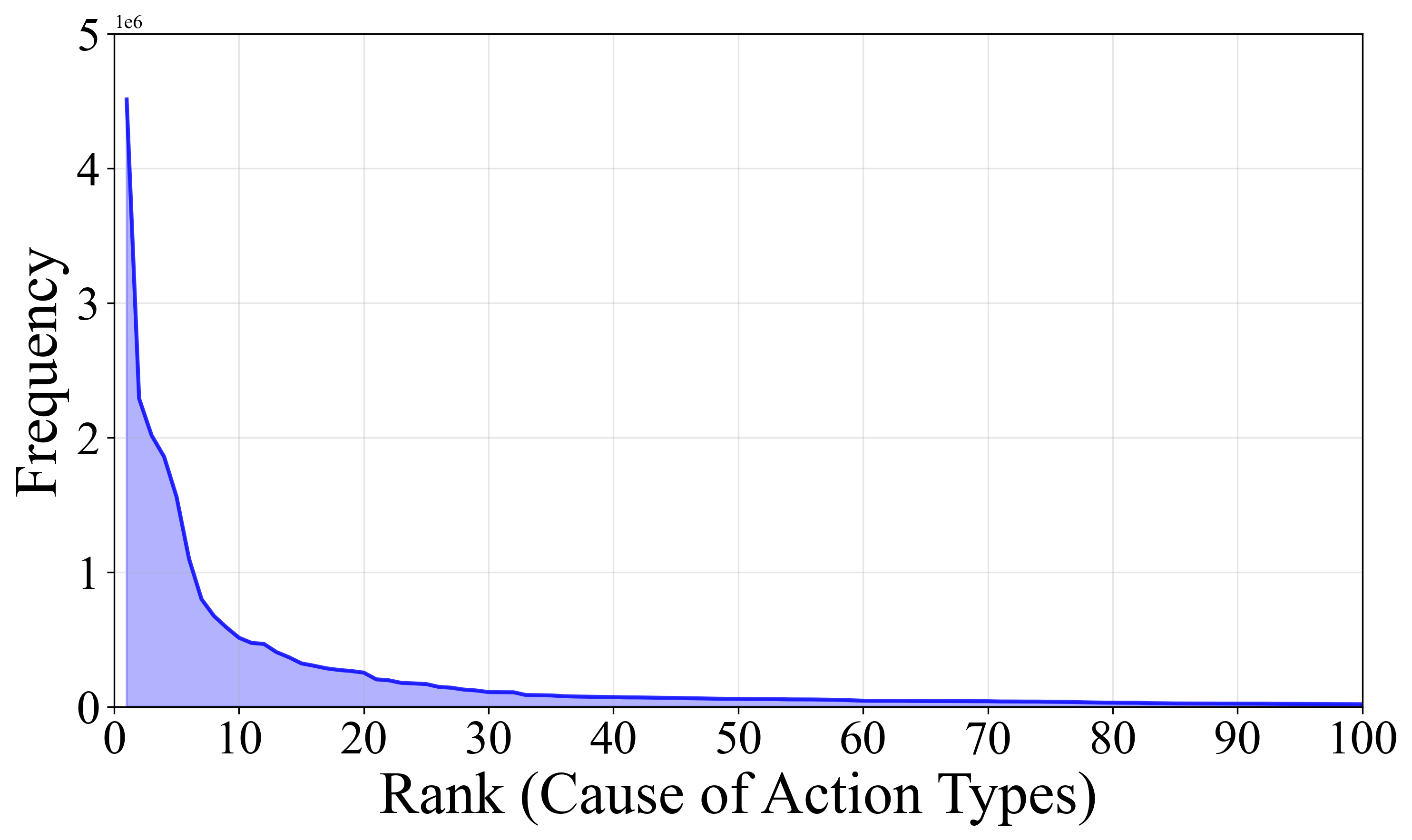}
    \caption*{(B) Distribution of top 100 most frequent causes of action in 26.6 million real-world civil court judgments.}
\end{minipage}

\caption{(A) and (B) show the distributions of top 100 most frequent legal relation types in \textbf{LexRel} and causes of action in real-world civil judgments. (A) shows a clear downward trend, which may reflect the long-tail pattern. (B) shows the frequency of the top cause types compared to those that appear less frequently, emphasizing the imbalance in occurrence and illustrating a clear long-tail pattern.}
\label{fig:longtail_two}
\end{figure*}

\subsubsection{Settings}
\paragraph{Models.} We select widely used state-of-the-art LLMs for evaluation, including closed-source models: GPT-4o \cite{hurst2024gpt}, o3-mini \cite{openai2025openai_o3mini} and Claude-Sonnet-4 \cite{anthropic2025claude_4}, as well as open-source models: DeepSeek-V3 \cite{deepseekai2025deepseekv3technicalreport}, DeepSeek-R1 \cite{deepseekai2025deepseekr1incentivizingreasoningcapability}\footnote{Although DeepSeek-R1 and DeepSeek-V3 are open-source models, we experimented via API calls due to computational resource considerations.}, Llama3.1-8B-Instruct, Llama3.1-70B-Instruct \cite{dubey2024llama}, InternLM3-8B-Instruct \cite{cai2024internlm2}, MiniCPM4-8B \cite{minicpmteam2025minicpm4ultraefficientllmsend}, Qwen3-8B, Qwen3-14B and Qwen3-32B \cite{yang2025qwen3}. 
Specifically, for relation-enhanced baseline, we select DeepSeek-R1 and GPT-4o as generators to produce the training data.

\paragraph{Configuration.} 
For the zero-shot setting, we directly evaluate the models on the \textbf{LexRel} benchmark. For the relation-enhanced setting, we first generate 5,500 synthetic training samples, and then fine-tune and evaluate the resulting models. Note that  due to training cost constraints, only the 8B and 14B variants of the open-source models are fine-tuned. 
Prompts used for both settings are presented in Table \ref{table2}. All experiments are conducted on 4$\times$A800 GPUs (each with 40GB memory).

\paragraph{Metrics.} We use precision, recall, micro-F1 score and macro-F1 score (computed by averaging F1 scores across different relation types) to assess model performance on both \textit{type extraction} and \textit{argument extraction} tasks.
For \textit{type extraction}, evaluation is conducted by directly matching each predicted legal relation type with the gold-standard label.
Considering the inherent ambiguity in legal language and the cost of large-scale human evaluation, we adopt LLM-as-a-Judge approach for evaluating \textit{argument extraction} correctness. Specifically, we pair each model-predicted argument with its corresponding gold argument and use DeepSeek-V3 to assess whether the two convey the same meaning.

To validate this choice, we randomly select 60 items, resulting in 193 pairs for manual inspection. A human expert independently judges whether each predicted–gold pair conveys matching information, and these expert decisions serve as the reference gold standard for evaluating the LLM-as-a-Judge strategy. Based on this analysis, we obtain a subject accuracy of $0.954$, an object accuracy of $0.969$, and a content accuracy of $0.810$. Taken together, these results show that DeepSeek-V3 is sufficiently reliable as an automatic evaluator in our setting.


\subsubsection{Results and Discussion}


The results of our evaluation are shown in Table \ref{table3}. 
In the zero-shot \textit{type extraction} task, the best overall performer is o3-mini (micro-F1 $= 0.762$), followed by DeepSeek-R1 (micro-F1 $= 0.693$). Among open-source models, models in the Qwen3 series demonstrate the strongest performance, with Qwen3-8B, Qwen3-14B and Qwen3-32B achieving micro-F1 scores of $0.464$, $0.583$ and $0.583$ respectively. These results suggest that reasoning LLMs are better suited for zero-shot \textit{type extraction}.
In contrast, \textit{argument extraction} proves significantly more challenging for the models. The highest zero-shot micro-F1 is again achieved by o3-mini $(0.382)$, followed by DeepSeek-R1 $(0.268)$ and Claude-Sonnet-4 $(0.258)$. Most open-source models perform poorly (micro-F1 $< 0.2$), indicating the difficulty of identifying argument spans in unstructured legal text under zero-shot conditions.

In the SFT experiments, we distill responses from GPT-4o and DeepSeek-R1 into smaller open-source models, and their performance improves significantly\footnote{In this paper, significance is tested through t-test ($\alpha=0.05$).} across both tasks. For \textit{type extraction}, the best distillation result is achieved by Qwen3-14B (micro-F1 $= 0.733$), which closely matches the performance of o3-mini in zero-shot baseline.
\textit{Argument extraction} also benefits significantly from SFT, for example, InternLM3-8B improves from $0.048$ to $0.323$ in micro-F1. This large gain suggests that SFT can effectively equip smaller models with the ability to learn structured legal relations.

Furthermore, we observe that the macro-F1 scores of all LLMs are lower than their corresponding micro-F1 scores across all settings. This disparity suggests that LLMs exhibit varying capabilities across different types of legal relations.

\subsection{Long-Tail Analysis}

\begin{figure}[t]
\centering
\includegraphics[width=\columnwidth]{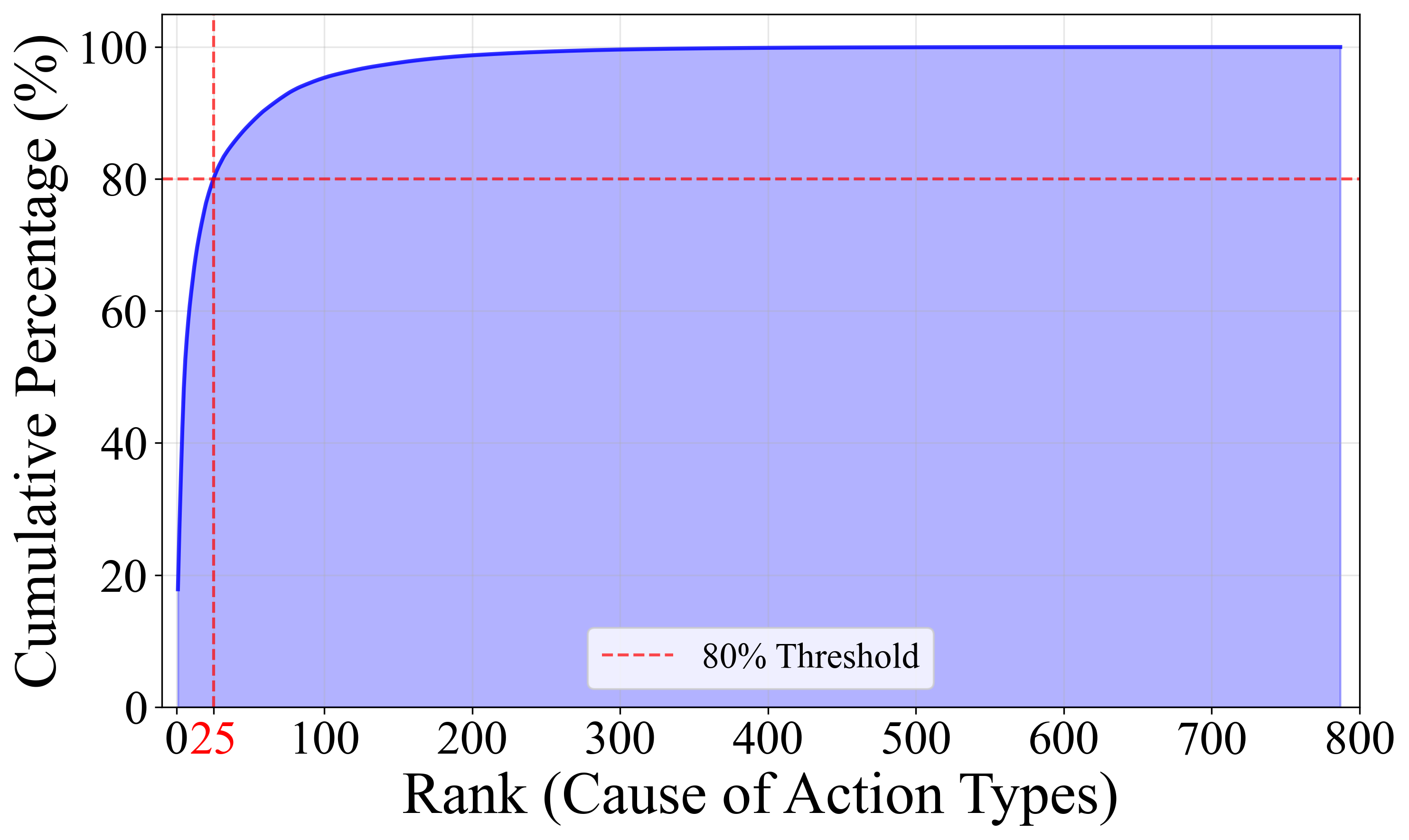} 
\caption{Pareto distribution of causes of action, demonstrating that the top 25 case types account for 80\% of all civil judgments. This graph highlights the dominance of a small number of case types in the overall dataset and identifies the 80\% threshold.
}
\label{fig:pareto}
\end{figure}


\textbf{LexRel} contains 1,140 civil cases and 1,863 fully annotated legal relations spanning all major civil law domains. 
Fig~\ref{fig:longtail_two}(A) shows the 100 most frequent legal relation types in \textbf{LexRel}, revealing a clear downward trend. We further examine whether this downward pattern faithfully mirrors real-world distributions. Because real-world cases do not explicitly specify legal relations, we use \textbf{civil cause of action} as a proxy. We first validate the reliability of this proxy by measuring the association between legal relations and causes of action in \textbf{LexRel}, which reveals a strong correlation
($\chi^2$ test, $p < 0.05$; Cramér’s V $= 0.528$)\footnote{Statistical significance and effect size are evaluated using a chi-square test and Cramér’s V. A result is conventionally regarded as a strong correlation when $p<0.05$ and $V>0.5$.}. 
We then examine a comprehensive set of 26.6 million real-world civil court judgments from China Judgments Online~\cite{cjo2013} and extract cause of action information for each case. Fig ~\ref{fig:longtail_two}(B) shows that causes of action for real-world civil cases follow a long-tail distribution. 
Consistently, Pareto analysis (Fig ~\ref{fig:pareto}) shows that a small number of frequent cause of action types($25$) account for the majority of instances, while a large number of types occur sparsely. 
Taken together, these results indicate that the long-tail distribution of legal relations in \textbf{LexRel} echoes the intrinsic pattern observed in real-world litigation data.

\subsection{Effects on Downstream Legal AI Tasks}
To demonstrate the effects of explicitly incorporating legal relations in legal AI tasks, we select a set of related tasks from LawBench \cite{fei2023lawbench} for further analysis. 

\subsubsection{Settings}
\paragraph{Tasks.} We select three tasks\textemdash Case Analysis, Consultation and Criminal Damages Calculation\textemdash from the LawBench dataset~\cite{fei2023lawbench}, based on their inherent dependency on legal relation identification. This dependency primarily manifests as a prerequisite step; for example, identifying the existence and type of legal relations prior to structured case analysis or legal consultation, or determining ownership or contractual relations as a basis for computing criminal damages.
\paragraph{Models.} We experiment with existing models, including GPT-4o, DeepSeek-V3, MiniCPM4-8B, and Qwen3-8B.
\paragraph{Hyperparameters.} Considering both efficiency and effectiveness, we use the legal relations distilled by DeepSeek-V3. Specifically, we extract relation types and arguments from the input and insert them directly into the task instructions during inference. All other evaluation parameters remain consistent with the previous section.
\paragraph{Metrics.} We follow the official evaluation guidelines of LawBench, where each task is scored on a scale from 0 to 100 based on the correctness and completeness of the model's output. We compare results under two settings: (1) vanilla zero-shot, where the model receives only the task instruction and input; and (2) relation-aware zero-shot, where the input is enhanced with distilled legal relation types and arguments.

\subsubsection{Results and Analysis}

\begin{table}[t]
  \centering
  \fontsize{8}{11}\selectfont
  \begin{tabularx}{\columnwidth}{X|cccc}
    \toprule
    \textbf{Models}  &\textbf{Case} & \textbf{Consultation} & \textbf{Calculation} \\
    \toprule
    \textbf{GPT-4o}          & 55.8 & 18.2 & 84.4 \\
    \textit{w/ LR}        & \textbf{56.6} & \textbf{19.2} & \textbf{85.8} \\
    \midrule
    \textbf{DeepSeek-V3}     & 66.2 & 16.4 & 85.0 \\
    \textit{w/ LR}    & \textbf{68.2} & \textbf{16.5} & \textbf{86.4} \\
    \midrule

    \textbf{MiniCPM4-8B}    & 32.0 & 7.6 & 65.0 \\
    \textit{w/ LR}    & \textbf{45.0} & \textbf{8.4} & \textbf{76.8} \\
    \midrule
    
    \textbf{Qwen3-8B}    & 57.2 & 15.0 & 91.8 \\
    \textit{w/ LR}    & \textbf{57.6} &14.2 & \textbf{93.2} \\

    \bottomrule
  \end{tabularx}
  \caption{Results on downstream legal AI tasks. The three tasks presented are: Case Analysis (Case), Consultation and Criminal Damages Calculation (Calculation). The notation `\textit{w/ LR}' refers to the setting where the model is evaluated using relation-aware inputs. Results which are improved by legal relation information are highlighted in \textbf{bold}.}
  \label{table4}
\end{table}

As shown in Table \ref{table4}, incorporating legal relations (``w/ LR'') consistently enhances performance across nearly all tasks and models. For Case Analysis and Consultation, legal relations provide essential cues for identifying and reasoning over relevant statutory knowledge. In Criminal Damages Calculation, certain types of legal relations (e.g., \textit{Legal Relation of Ownership}) enable the model to more accurately identify property that has been subject to infringement, thereby enabling the calculation of damages. 
Although current LLMs show limited performance in extracting legal relations, incorporating these relations into downstream tasks still yields performance benefits, suggesting that legal relation modeling remains a promising direction for further exploration.

\section{Conclusion}
In this work, we explicitly model legal relations in the context of Chinese civil law. We introduce a structured schema of civil legal relations and curate \textbf{LexRel}, a benchmark for legal relation extraction. Based on this benchmark, we evaluate large language models on their ability to extract legal relations and associated arguments, and demonstrate that integrating such structured knowledge improves performance across multiple downstream legal AI tasks.

\section{Limitations}
Our schema and dataset focus specifically on civil legal relations under Chinese law, which may limit direct transferability to other jurisdictions. Although the conceptual structure of legal relations is broadly consistent across civil law systems, each jurisdiction embodies distinct doctrines and terminology, requiring localized adaptation. 

A second limitation concerns potential model-family coupling. Both the generation of synthetic data used for supervised fine-tuning (SFT) and the LLM-as-a-Judge evaluation involve models from the DeepSeek family. Although we adopt mitigation strategies to reduce stylistic bias during evaluation, a perceived dependency across data generation, training and evaluation may still remain. Future work could address this by diversifying model sources and further refining evaluation protocols.


\section{Ethical Considerations}
All data used in this study are sourced from publicly available Chinese civil court decisions released through official judicial channels~\cite{cjo2013}. We strictly adhere to data privacy and research ethics standards, ensuring that no confidential information is disclosed. All human annotators participated voluntarily and are compensated fairly according to institutional guidelines.

\section{Acknowledgments}

This work was supported by the Tsinghua University Initiative Scientific Research Program (20255080016). We sincerely thank all annotators for their meticulous efforts in refining the \textbf{LexRel} benchmark, the legal experts for their invaluable guidance on the taxonomy design and annotation process, and the anonymous reviewers for their constructive feedback and suggestions.



\bibliography{custom}

\appendix

\section{Prompts}
\label{appendix:c}
The complete definition of arguments and the prompts used in \textit{argument extraction} are listed in Tables \ref{table5}-\ref{table8}. 

\section{Inter-annotator Agreement}
\label{appendix:d}
Because evaluation of \textit{argument extraction} in \textit{legal relation extraction} involves semantic alignment between generated textual arguments rather than exact label matching, inter-annotator agreement (IAA) is assessed based on expert judgment. Accordingly, a senior legal expert is designated as the reference annotator for agreement assessment.

Owing to the substantial cost of fully annotating and validating the entire dataset, we randomly sample 300 instances for agreement assessment. These instances are evenly distributed among other annotators for independent annotation. The senior expert subsequently reviews each annotation by comparing it with a reference interpretation derived from the same annotation guidelines. When discrepancies arise, the instance is adjudicated by the senior expert to establish a new reference annotation. IAA is then computed between the senior expert and the other annotators, yielding a Cohen’s Kappa~\cite{cohen1960coefficient} of $0.706$.\footnote{According to~\citet{landis1977measurement}, a Cohen's Kappa score between 0.61 and 0.80 signifies ``substantial agreement''.}

\section{Error Analysis}
\label{sec:error_analysis}

To better understand the limitations of LLMs in legal relation extraction, we conduct a qualitative analysis of representative failure cases.
We find that the errors mainly arise from two distinct sources: (1) systematic biases in type prediction, which lead to over-generalization and poor coverage of rare categories, and (2) insufficient granularity in argument extraction, which results in incomplete or inaccurate identification of legally critical elements.

\begin{figure*}[t]
\centering
\includegraphics[width=\textwidth, keepaspectratio]{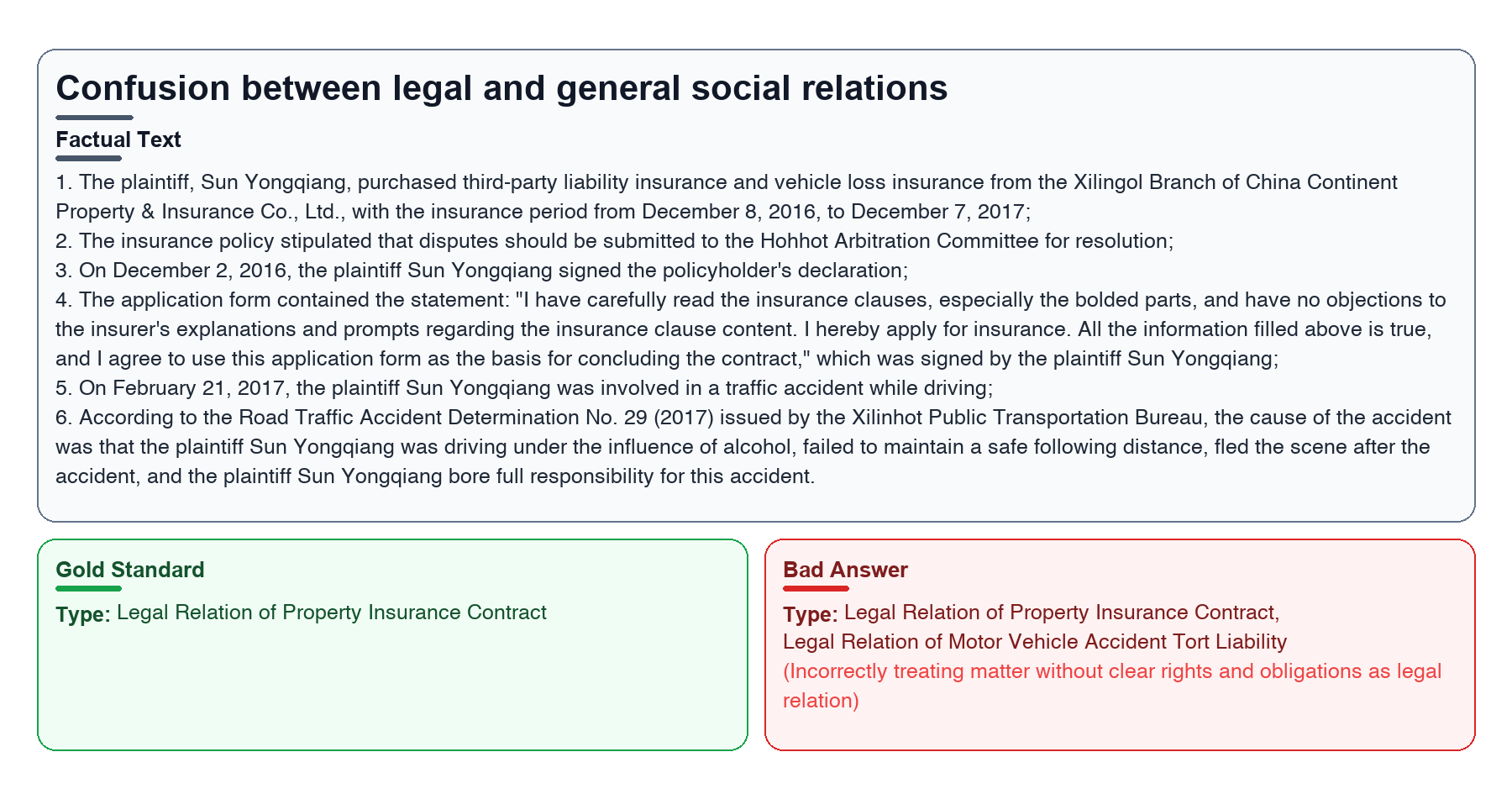} 
\caption{An example of confusion error, where the model correctly identifies the property insurance contract relation but incorrectly predicts an additional motor vehicle accident tort relation.}
\label{fig:row1}
\end{figure*}

\begin{figure*}[ht]
\centering

\includegraphics[width=\textwidth, keepaspectratio]{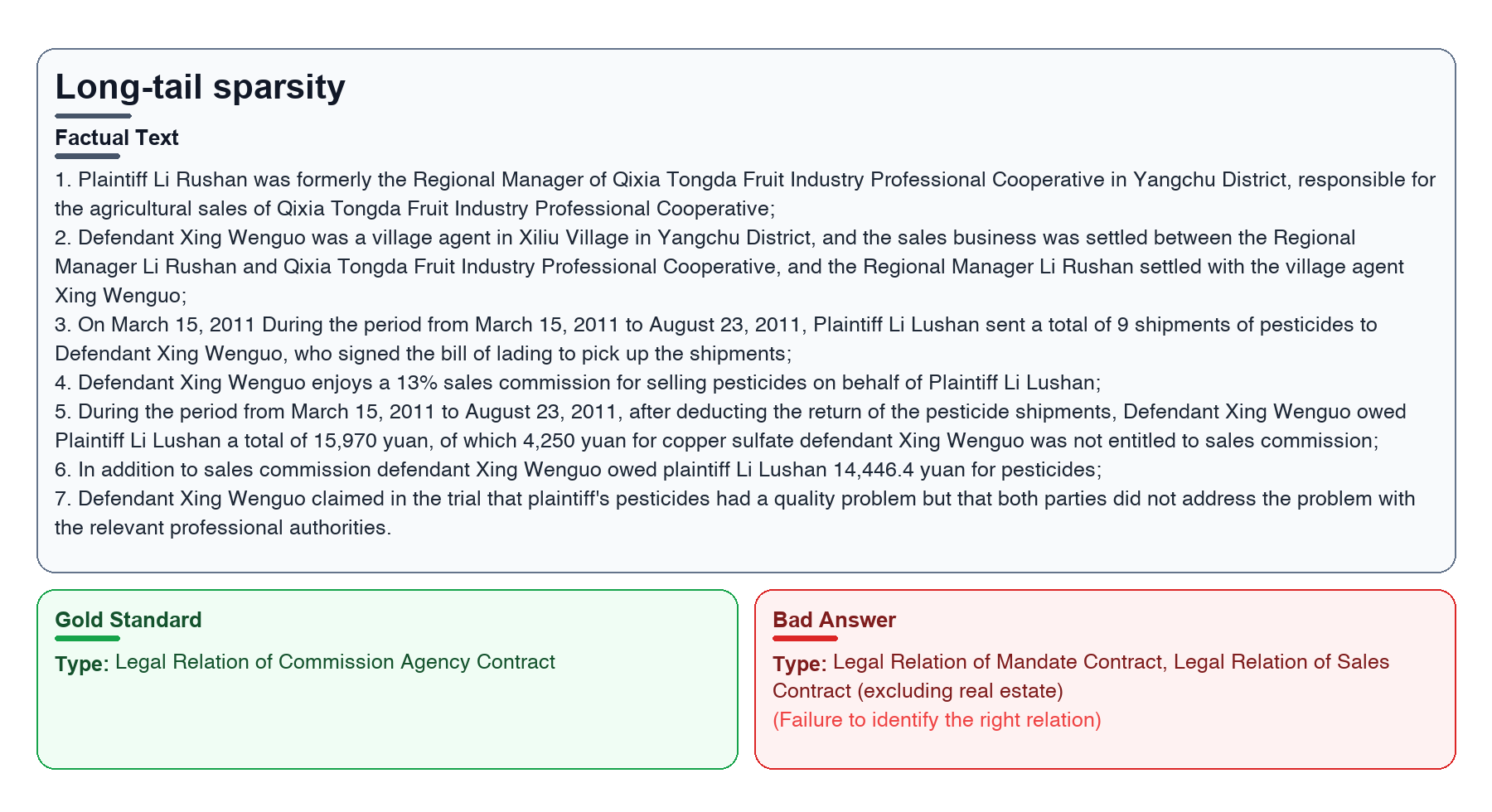} 
\caption{An example of long-tail sparsity error, where the model fails to identify the commission agency contract and instead maps the case to more frequent relation types.}
\label{fig:row2}
\end{figure*}

\paragraph{Type extraction errors.} Errors in predicting legal relation types primarily stem from two factors, leading to both over-prediction of common relations and poor recognition of rare ones:
\begin{itemize}
    \item \textbf{Confusion between legal and general social relations:}
    As shown in Fig \ref{fig:row1}, LLMs tend to over-generalize based on surface-level factual semantics, frequently mapping ordinary social or factual relations—without legally binding rights and obligations—into formal legal relations.
    This bias causes systematic over-prediction, resulting in inflated recall but significantly reduced precision.
    Taking zero-shot results as an illustrative example, this issue occurs in $81.1\%$ of the predictions of Llama3.1-8B-Instruct, which exhibits relatively weak performance, while even the strongest model in our evaluation, o3-mini, still exhibits this error in $28.6\%$ of its predictions. 
    \item \textbf{Long-tail sparsity:}  
    As shown in Fig \ref{fig:longtail_two}, legal relation types in real-world cases follow a highly imbalanced distribution. For rare or unseen relation types, LLMs lack sufficient exposure during training and therefore fail to learn robust representations.  
    As a result, as shown in Fig \ref{fig:row2}, they often misclassify these instances into high-frequency categories, leading to poor performance on tail classes and reduced diversity in predicted relation types.
\end{itemize}

\begin{figure*}[h!]
\centering
\includegraphics[width=\textwidth]{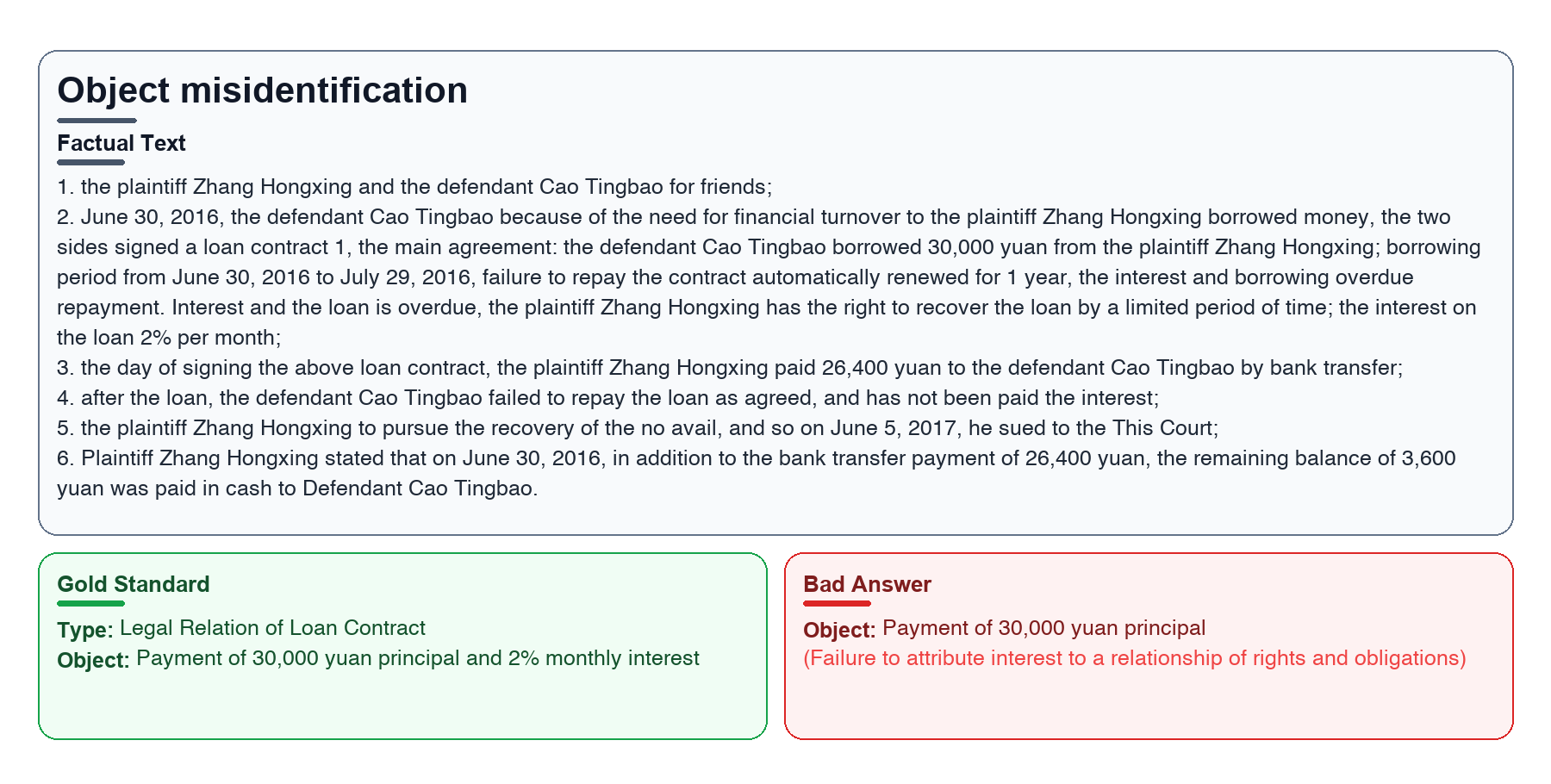} 
\caption{An example of object misidentification, where the model extracts only the principal payment while omitting the interest component in the contractual object.}
\label{fig:row3}
\end{figure*}

\begin{figure*}[h!]
\centering
\includegraphics[width=\textwidth]{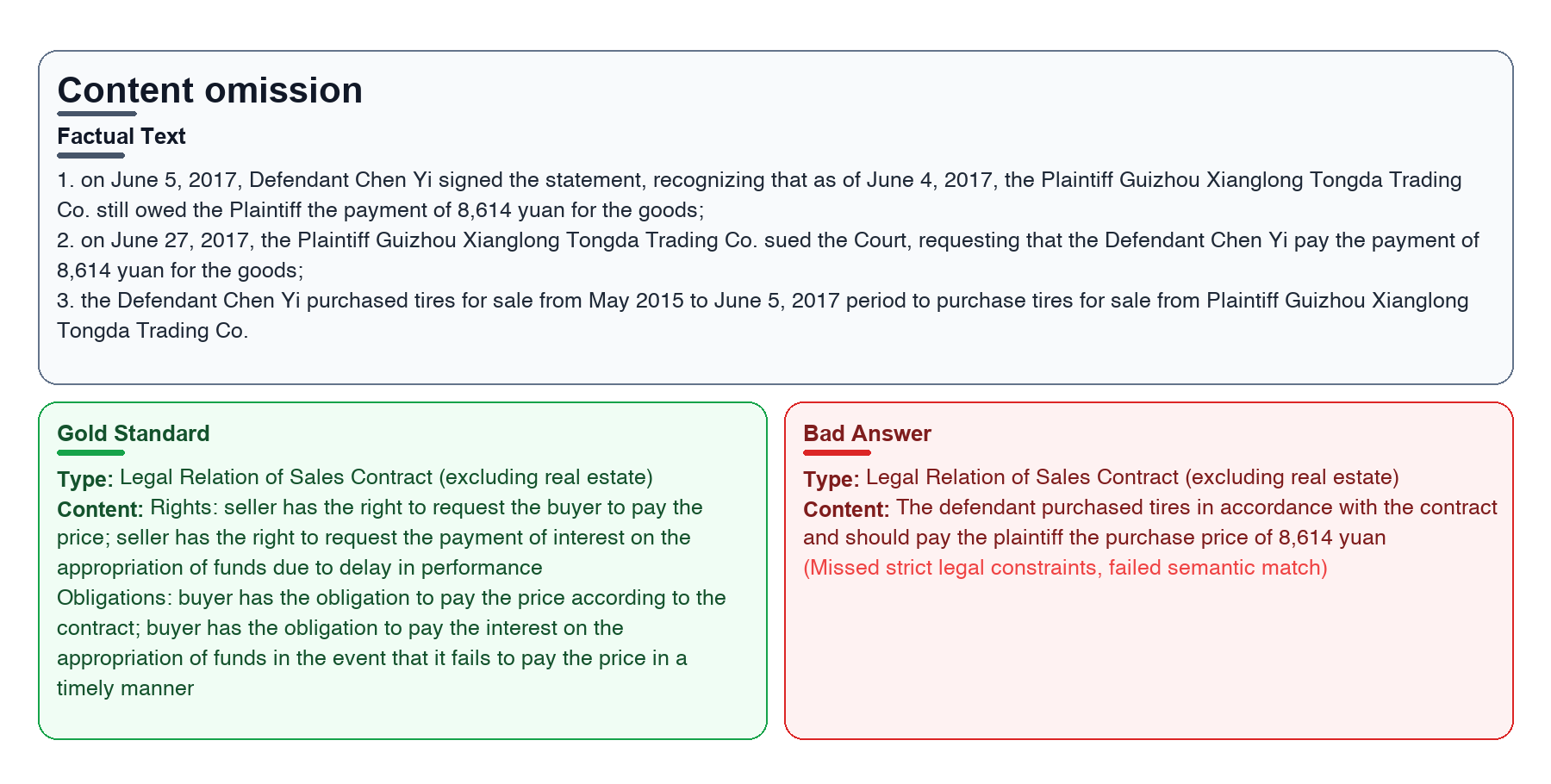} 
\caption{An example of content omission, where the model captures the sales contract relation but fails to recover the complete fine-grained rights and obligations in the legal content.}
\label{fig:row4}
\end{figure*}

\paragraph{Argument extraction errors.}
In the argument extraction stage, errors mainly arise from the model's inability to capture fine-grained legal elements, leading to incomplete or inaccurate reconstruction of the legal relation:
\begin{itemize}
    \item \textbf{Object misidentification:}
    LLMs frequently fail to correctly identify the object of a legal relation, especially when a case involves multiple parties or entities.
    As shown in Fig \ref{fig:row3}, models struggle to identify the object that is truly relevant to the underlying rights–obligations relationship, and instead default to objects associated with general, non-legal relations.
    \item \textbf{Content omission:}  
    The extraction of content—the specific rights and obligations—is often incomplete.  
    As shown in Fig \ref{fig:row4}, LLMs tend to generate partial summaries that omit key legally binding elements. Under strict semantic matching evaluation, these missing components lead to penalties, as the predicted content fails to fully align with the ground truth.
\end{itemize}

\begin{table*}[h]
\centering
\fontsize{9}{11}\selectfont
\begin{tabularx}{\textwidth}{c|X}
    \toprule
    \textbf{Types} & \textbf{Prompt} \\
    \toprule
    Legal Relations of Personality Rights &
    Please extract the subject, object, and content of \textbf{\{relation type\}}  from the given judgment documents without explanation or description.

    \textbf{Factual text: \{\}}.
    
    \textbf{Subject definition}: The right holder himself/herself and all other subjects are the obligors (generally only the right holder is output).
    
    \textbf{Object definition}: The corresponding personal interests (output xx rights interests only).
    
    \textbf{Content definition}: In terms of rights, the subject enjoys the right to have their life, body, health, name, portrait, reputation, privacy and other personal interests not infringed upon by law. In terms of obligations, other civil subjects bear negative obligations such as not infringing upon, not insulting or defaming, and not disclosing relevant information. When necessary, they also have the obligation to actively protect and provide relief.
    
    Output requirements:
    
    1. Each set of results is presented as a standard JSON object.
    
    2. Required fields: \{`subject': `', `object': `', `content': `'\}.
    
    3. Each set is in a separate paragraph without numbering or sorting.
    
    4. Only one set needs to be output for the same subject and object. \\
    \midrule

    Legal Relations of Status Rights &
    Please extract the subject, object, and content of \textbf{\{relation type\}}  from the given judgment documents without explanation or description.

    \textbf{Factual text: \{\}}.
    
    \textbf{Subject definition}: The right holder themselves and all other subjects are the obligors (generally, multiple right holders and all obligors related to the facts should be output simultaneously).
    
    \textbf{Object definition}: Corresponding identity interests (output xx relationship interests only).
    
    \textbf{Content definition}: In terms of rights, the subject enjoys the rights of spousal, parental, guardianship, support and inheritance, etc. in accordance with the law due to the identity relationship. In terms of obligations, the corresponding subjects undertake identity obligations such as loyalty, assistance, support, education, and guardianship.
    
    Output requirements:
    
    1. Each set of results is presented as a standard JSON object.
    
    2. Required fields: \{`subject': `', `object': `', `content': `'\}.
    
    3. Each set is in a separate paragraph without numbering or sorting.
    
    4. Only one set needs to be output for the same subject and object. \\
    \midrule

    Tortious Legal Relations &
    Please extract the subject, object, and content of \textbf{\{relation type\}}  from the given judgment documents without explanation or description.

    \textbf{Factual text: \{\}}.
    
    \textbf{Subject definition}: The infringer is the subject that implements the infringement act, and the infringed party is the subject that suffers damage due to the infringement act (generally, both should be output simultaneously).
    
    \textbf{Object definition}: The legitimate rights and interests of the infringed party (just output the name of the infringed right).
    
    \textbf{Content definition}: The rights enjoyed by the creditor: The right to request the debtor to perform or refrain from acting in accordance with the agreement or law. When a debtor fails to exercise his rights or infringes upon his creditor's rights, the right of subrogation or revocation may be exercised in accordance with the law. The obligations borne by the debtor: They should fulfill the obligation of payment as agreed, including delivering money, property or providing services, and may also bear the obligation of inaction.
    
    Output requirements:
    
    1. Each set of results is presented as a standard JSON object.
    
    2. Required fields: \{`subject': `', `object': `', `content': `'\}.
    
    3. Each set is in a separate paragraph without numbering or sorting.
    
    4. Only one set needs to be output for the same subject and object. \\
    
    \bottomrule
\end{tabularx}
\caption{Definitions and prompt templates used in \textit{argument extraction} for \textit{Legal Relations of Personality Rights}, \textit{Legal Relations of Status Rights} and \textit{Tortious Legal Relations} are shown (English translation for the original Chinese text).}
\label{table5}
\end{table*}

\begin{table*}[h]
\centering
\fontsize{9}{11}\selectfont
\begin{tabularx}{\textwidth}{c|X}
    \toprule
    \textbf{Types} & \textbf{Prompt} \\
    \toprule

    Legal Relations of Intellectual Property &
    Please extract the subject, and object of \textbf{\{relation type\}}  from the given judgment documents without explanation or description.

    \textbf{Factual text: \{\}}.
    
    \textbf{Subject definition}: The right holder himself/herself and all other subjects are the obligors (generally only the right holder is output).
    
    \textbf{Object definition}: Intellectual achievements, including works, inventions and creations, trademarks, trade secrets, etc. (Output the name of the intellectual achievement and the corresponding rights to this achievement).
    
    Output requirements:
    
    1. Each set of results is presented as a standard JSON object.
    
    2. Required fields: \{`subject': `', `object': `'\}.
    
    3. Each set is in a separate paragraph without numbering or sorting.
    
    4. Only one set needs to be output for the same subject and object. \\
    \midrule

    Legal Relations of Real Rights &
    Please extract the subject, object, and content of \textbf{\{relation type\}}  from the given judgment documents without explanation or description.

    \textbf{Factual text: \{\}}.
    
    \textbf{Subject definition}: The right holder himself/herself and all subjects related to the rights and obligations relationship are the obligors (generally the output right holder and other obligors who bear direct legal obligations in this fact).
    
    \textbf{Object definition}: Tangible objects, including immovable property, movable property and special objects. (Name of the output).

    \textbf{Content definition}: In terms of rights, the right holder enjoys ownership, usufructuary rights, security interests, etc. in accordance with the law, exercises direct control over the object through rights such as possession, use, income, and disposal, and can legally exclude interference from others. Immovable property rights come into effect through registration, while movable property rights take delivery as a prerequisite for their effectiveness. In terms of obligations, it includes completing the registration or delivery procedures in accordance with the law; It is prohibited to infringe upon the legitimate property rights of others. When property rights are illegally interfered with, it undertakes the obligation of protection such as removing obstacles, restoring the original state and returning the original property.
    
    Output requirements:
    
    1. Each set of results is presented as a standard JSON object.
    
    2. Required fields: \{`subject': `', `object': `', `content': `'\}.
    
    3. Each set is in a separate paragraph without numbering or sorting.
    
    4. Only one set needs to be output for the same subject and object. \\

    \midrule
    
    Legal Relations of Success &
    Please extract the subject, object, and content of \textbf{\{relation type\}}  from the given judgment documents without explanation or description.

    \textbf{Factual text: \{\}}.
    
    \textbf{Subject definition}: Heir and decedent. (Generally, both need to be output simultaneously).

    \textbf{Object definition}: Specific heritage content.
    
    \textbf{Content definition}: In terms of rights, heirs enjoy rights such as inheriting the estate, accepting bequests, and subrogation inheritance in accordance with the law or will. In terms of obligations, the heirs should divide the property in the spirit of friendly consultation, fulfill the obligations stipulated in the conditional will, notify the relevant parties and properly keep the estate, and also settle debts and taxes in accordance with the law.
    
    Output requirements:
    
    1. Each set of results is presented as a standard JSON object.
    
    2. Required fields: \{`subject': `', `object': `', `content': `'\}.
    
    3. Each set is in a separate paragraph without numbering or sorting.
    
    4. Only one set needs to be output for the same subject and object. \\
    
    \bottomrule
\end{tabularx}
\caption{Definitions and prompt templates used in \textit{argument extraction} for \textit{Legal Relations of Intellectual Property}, \textit{Legal Relations of Real Rights} and \textit{Legal Relations of Success} are shown (English translation for the original Chinese text).}
\label{table6}
\end{table*}

\begin{table*}[h]
\centering
\fontsize{8}{10}\selectfont
\begin{tabularx}{\textwidth}{c|X}
    \toprule
    \textbf{Types} & \textbf{Prompt} \\
    \toprule

    Contractual Legal Relations &
    Please extract the subject, object, and content of \textbf{\{relation type\}}  from the given judgment documents without explanation or description.

    \textbf{Factual text: \{\}}.
    
    \textbf{Subject definition}: The parties to the contract (generally all parties need to be output simultaneously).
    
    \textbf{Object definition}: The subject matter pointed to by the contractual obligations, including goods, acts, rights, etc. (output the performance of the contract).
    
    \textbf{Content definition}: The rights enjoyed by the creditor: The right to request the debtor to perform or refrain from acting in accordance with the agreement or law. When a debtor fails to exercise his rights or infringes upon his creditor's rights, the right of subrogation or revocation may be exercised in accordance with the law. The obligations borne by the debtor: They should fulfill the obligation of payment as agreed, including delivering money, property or providing services, and may also bear the obligation of inaction.
    
    Output requirements:
    
    1. Each set of results is presented as a standard JSON object.
    
    2. Required fields: \{`subject': `', `object': `', `content': `'\}.
    
    3. Each set is in a separate paragraph without numbering or sorting.
    
    4. Only one set needs to be output for the same subject and object. \\
    \midrule

    Legal Relation of Unjust Enrichment &
    Please extract the subject, object, and content of \textbf{\{relation type\}}  from the given judgment documents without explanation or description.

    \textbf{Factual text: \{\}}.
    
    \textbf{Subject definition}: A beneficiary is the subject that gains benefits without legal basis, while a victim is the subject that suffers losses due to the beneficiary's acquisition of benefits (generally, both should be output simultaneously).
    
    \textbf{Object definition}: Interest, which refers to the property or benefit obtained by the beneficiary without legal basis. (Content of Output Benefits).
    
    \textbf{Content definition}: The rights enjoyed by the creditor: The right to request the debtor to perform or refrain from acting in accordance with the agreement or law; When a debtor fails to exercise his rights or infringes upon his creditor's rights, the right of subrogation or revocation may be exercised in accordance with the law. The obligations borne by the debtor: They should fulfill the obligation of payment as agreed, including delivering money, property or providing services, and may also bear the obligation of inaction.
    
    Output requirements:
    
    1. Each set of results is presented as a standard JSON object.
    
    2. Required fields: \{`subject': `', `object': `', `content': `'\}.
    
    3. Each set is in a separate paragraph without numbering or sorting.
    
    4. Only one set needs to be output for the same subject and object. \\
    \midrule

    Legal Relation of Negotiorum Gestio &
    Please extract the subject, object, and content of \textbf{\{relation type\}}  from the given judgment documents without explanation or description.

    \textbf{Factual text: \{\}}.
    
    \textbf{Subject definition}: A manager (creditor) refers to a person who, without being entrusted or bound by legal obligations, takes the initiative to manage the affairs of others to prevent their interests from being damaged. The beneficiary (debtor) refers to the party whose interests are attributed to the management of the affairs (usually both of them should be output simultaneously).
    
    \textbf{Object definition}: The actions of the manager (such as repairing houses on behalf of others, providing assistance to the sick and injured, etc.) and the necessary expenses or compensation for losses arising therefrom. (Content of the Output act).
    
    \textbf{Content definition}: The rights enjoyed by the creditor: The right to request the debtor to perform the act or refrain from acting in accordance with the agreement or law; When a debtor fails to exercise his rights or infringes upon his creditor's rights, the right of subrogation or revocation may be exercised in accordance with the law. The obligations borne by the debtor: They should fulfill the obligation of payment as agreed, including delivering money, property or providing services, and may also bear the obligation of inaction.
    
    Output requirements:
    
    1. Each set of results is presented as a standard JSON object.
    
    2. Required fields: \{`subject': `', `object': `', `content': `'\}.
    
    3. Each set is in a separate paragraph without numbering or sorting.
    
    4. Only one set needs to be output for the same subject and object. \\

    \bottomrule
\end{tabularx}
\caption{Definitions and prompt templates used in \textit{argument extraction} for \textit{Contractual Legal Relations}, \textit{Legal Relation of Unjust Enrichment} and \textit{Legal Relation of Negotiorum Gestio} are shown (English translation for the original Chinese text).}
\label{table7}
\end{table*}

\begin{table*}[h]
\centering
\fontsize{8}{9}\selectfont
\begin{tabularx}{\textwidth}{c|X}
    \toprule
    \textbf{Types} & \textbf{Prompt} \\
    \toprule

    Legal Relation of Letters of Credit &
    Please extract the subject, object, and content of \textbf{\{relation type\}}  from the given judgment documents without explanation or description.

    \textbf{Factual text: \{\}}.
    
    \textbf{Subject definition}: The primary parties are the Applicant, the Issuing Bank, and the Beneficiary, but there are eight parties in total, including: the Advising Bank, the Confirming Bank, the Negotiating Bank, the Paying Bank, and the Reimbursing Bank.

    \textbf{Object definition}: The delivery of documents and the promise of payment.
    
    \textbf{Content definition}:  
    1. The contractual sales relationship between the Applicant and the Beneficiary.  
    2. The principal-agent contractual relationship between the Issuing Bank and the Applicant.  
    3. The relationship between the Issuing Bank and the Beneficiary (a subject of theoretical and practical debate). Generally, when a revocable letter of credit is issued, there is no contractual relationship between the Issuing Bank and the Beneficiary. However, when an irrevocable letter of credit is issued and received by the Beneficiary, a binding independent contractual relationship is formed between them.  
    4. The principal-agent relationship between the Advising Bank and the Issuing Bank.
    5. The contractual relationship between the Issuing Bank and the Paying/Negotiating Bank arises when the Issuing Bank designates or authorizes another bank to pay or negotiate, and that bank accepts the role. Otherwise, no legal relationship exists. 
    6. Typically, there is no legal relationship between the Beneficiary and the Paying/Negotiating Bank, unless a bank (other than the Issuing Bank) pays or negotiates the Beneficiary’s draft under the Issuing Bank’s authorization. In such cases, a creditor-debtor legal relationship is established under negotiable instruments law.
    
    Output requirements:
    
    1. Each set of results is presented as a standard JSON object.
    
    2. Required fields: \{`subject': `', `object': `', `content': `'\}.
    
    3. Each set is in a separate paragraph without numbering or sorting.
    
    4. Only one set needs to be output for the same subject and object. \\

    \midrule

    Legal Relation of Independent Guarantees &
    Please extract the subject, object, and content of \textbf{\{relation type\}}  from the given judgment documents without explanation or description.

    \textbf{Factual text: \{\}}.
    
    \textbf{Subject definition}: Issuer (bank/financial institution), beneficiary, applicant (can be an instructor).
    
    \textbf{Object definition}: The "document" on which the claim request is based.
    
    \textbf{Content definition}: It refers to the bill rights that the subject enjoys and the bill obligations it should undertake in accordance with the law. The right to a bill is the right enjoyed by the right holder to request the obligor to pay the amount of the bill, including the right to request payment and the right of recourse. Bill obligations refer to the responsibilities that the obligor must fulfill to satisfy the rights claims enjoyed by the right holder based on the bill, such as payment obligations, acceptance obligations, and guarantee payment obligations, etc.
    
    Output requirements:
    
    1. Each set of results is presented as a standard JSON object.
    
    2. Required fields: \{`subject': `', `object': `', `content': `'\}.
    
    3. Each set is in a separate paragraph without numbering or sorting.
    
    4. Only one set needs to be output for the same subject and object. \\

    \midrule

    Legal Relation of Bills &
    Please extract the subject, object, and content of \textbf{\{relation type\}}  from the given judgment documents without explanation or description.

    \textbf{Factual text: \{\}}.
    
    \textbf{Subject definition}: A subject is specific and mainly includes: the drawer, the payee, the payer, the holder of the bill, the acceptor, the endorser, the guarantor, the participant, etc.
    
    \textbf{Object definition}: A certain amount of currency.
    
    \textbf{Content definition}: The "underlying transaction contract" relationship between the applicant and the beneficiary; The "independent guarantee contract" relationship between the issuer and the beneficiary; The "application/recovery contract" relationship between the issuer and the applicant.
    considerations correspond to: the issuer's unconditional payment obligation, the principle of document review and "apparent conformity", the right of recovery after payment, the right of defense against fraud exceptions, and the conditions for termination or invalidation of the guarantee.
    
    Output requirements:
    
    1. Each set of results is presented as a standard JSON object.
    
    2. Required fields: \{`subject': `', `object': `', `content': `'\}.
    
    3. Each set is in a separate paragraph without numbering or sorting.
    
    4. Only one set needs to be output for the same subject and object. \\
    
    \bottomrule

\end{tabularx}
\caption{Definitions and prompt templates used in \textit{argument extraction} for \textit{Legal Relation of Letters of Credit}, \textit{Legal Relation of Independent Guarantees} and \textit{Legal Relation of Bills} are shown (English translation for the original Chinese text).}
\label{table8}
\end{table*}

\end{document}